\documentclass[letterpaper, 10 pt, conference]{ieeeconf}
\usepackage{cite} 
\usepackage{multirow}
\usepackage[utf8]{inputenc}
\usepackage[misc]{ifsym}
\usepackage{graphicx}
\usepackage{tikz}
\usepackage{pifont}

\usepackage{bbding}
\usepackage{pdfpages}
\usepackage{amsmath}
\usetikzlibrary{arrows, shapes, positioning, shadows, trees, calc, decorations.markings}

\usepackage{amssymb}
\usepackage{hyperref}
\usepackage{booktabs}
\usepackage{algorithm}
\usepackage{algpseudocode}
\usepackage{balance}
\usepackage{float}
\usepackage{color, colortbl}
\definecolor{LightCyan}{rgb}{0.88,1,1}
\makeatletter
\renewcommand\fs@ruled{%
  \def\@fs@cfont{\bfseries}%
  \let\@fs@capt\floatc@ruled
  \def\@fs@pre{\kern6pt\hrule height.8pt depth0pt \kern2pt}%
  \def\@fs@post{\kern2pt\hrule\relax}%
  \def\@fs@mid{\kern2pt\hrule\kern2pt}%
  \let\@fs@iftopcapt\iftrue
}
\makeatother
\hypersetup{
colorlinks=true,
linkcolor=black
}

\makeatletter
\newcommand{\printfnsymbol}[1]{%
  \textsuperscript{\@fnsymbol{#1}}%
}

\makeatother

\IEEEoverridecommandlockouts                

\title{\LARGE \bf
Trajectory Divergence Horizon Decision for Reliable Dual-Arm Surgical Subtask Manipulation
}

\author{Mingwu Su$^{1}$\printfnsymbol{1}\thanks{\printfnsymbol{1} Equal contribution},
Guankun Wang$^{1}$\printfnsymbol{1}, Jinsong Lin$^{1}$\printfnsymbol{1}, Rulin Zhou$^{1,3}$, Ziyi Hao$^{1}$, Zhiwei Fang$^{1}$, 
\\ Huxin Gao$^{1}$, Jiewen Lai$^{1}$, Jiazheng Wang$^{2}$, Fan Zhang$^{2}$, Hongliang Ren$^{1,3,\dagger}$, \textit{Senior Member, IEEE}
\thanks{$^{1}$Department of Electronic Engineering, 
The Chinese University of Hong Kong (CUHK), Hong Kong, China.} \thanks{$^{2}$ The Theory Lab, Central Research Institute, 2012 Labs, Huawei Technologies Co. Ltd., Hong Kong SAR, China.}
\thanks{$^{3}$ Shenzhen Loop Area Institute, Shenzhen, China.}
\thanks{
The work was supported by Ministry of Science and Technology (MOST) of China Key Project 2025YFE0122500, 2024YFE0216200, Innovation and Technology Fund (ITF) of Hong Kong SAR (ITF MHP/185/24), CUHK Direct Grant for Research 2024/2025 (4055262), Young Talent Support Project of Guangzhou Association for Science and Technology (QT-2025-046), and Hong Kong Research Grants Council (Grant No.: 14200425). (Corresponding author: Hongliang Ren.)
}
}

\begin{document}
\maketitle
\thispagestyle{empty}
\pagestyle{empty}

\begin{abstract}
Surgical robotic systems are increasingly being adopted as clinical workload rises, motivating autonomous solutions for repetitive manipulation subtasks. Learning-based controllers improve generalization compared with rule-based and analytic approaches, but most are trained for individual tasks and remain difficult to reuse across procedures. Vision-Language-Action (VLA) models provide a unified framework that integrates visual perception, language grounding, and action generation, offering a promising path toward more composable surgical autonomy. However, existing VLA policies rely on fixed-length open-loop action sequences, where changing scene conditions can lead to accumulated errors and potential risks in surgical manipulation. To mitigate this issue, we formulate surgical VLA deployment as an adaptive execution-horizon decision problem and propose Trajectory Divergence Horizon Decision (TDHD), a test-time mechanism that estimates step-wise action reliability by measuring the divergence between two flow-matching-generated trajectories under small noise perturbations and truncates execution using a dual-threshold rule to trigger timely replanning. We further establish a real-world da Vinci-like dual-arm benchmark with synchronized multi-view perception and language instructions, and collect 600 teleoperated demonstrations across needle (reach, pick, regrasp) and tissue (reach, lift, resection) manipulation suites. On real hardware with 20 trials per task setting, TDHD consistently improves performance over the latest VLA baselines: success increases from 55\% to 60\% for needle manipulation and from 55\% to 80\% for tissue manipulation, with the largest gains observed in the final manipulation stages. These results highlight the importance of adaptive execution control for reliable deployment of VLA models in surgical robotic manipulation.
\end{abstract}

\section{Introduction}

Surgical robotic systems have enabled millions of procedures worldwide and become a foundational infrastructure in modern operative care~\cite{dupont2021decade,long2025surgical}. With rapid population aging and increasing procedural demand, clinical workload continues to rise, placing sustained pressure on surgeons. This trend motivates surgical autonomy as a practical strategy to alleviate human burden, particularly for subtasks characterized by repetitive execution and well-defined procedural structure, where autonomy can enhance execution consistency~\cite{saeidi2022autonomous}. 

Traditional approaches to automated surgical robotics predominantly rely on rule-based design, geometric constraints, or predefined analytical models, and are often tailored to specific instruments and narrowly scoped scenarios~\cite{murali2015learning}. However, surgical environments are inherently complex, involving deformable soft tissue, evolving scene context, and multi-stage workflows that are difficult to model explicitly~\cite{liu2026real}. Data-driven learning methods offer an alternative paradigm by encoding diverse manipulation skills within a unified framework and improving generalization through task composition and data scaling~\cite{zhang2025deep}. Nevertheless, most learning-based surgical controllers are still developed in a task-by-task manner, where separate policies are trained and deployed for different subtasks, limiting cross-task reuse and composability at the system level. Recent advances in embodied intelligence, particularly Vision-Language-Action (VLA) models, demonstrate strong capability in general-purpose robotic manipulation~\cite{din2025vision}. Conditioned on visual observations and language instructions, these models generate action sequences or temporally extended action chunks, thereby integrating perception, semantic grounding, planning, and control within a single architecture~\cite{schmidgall2024general}. This paradigm establishes a foundation for learning-based surgical autonomy with reduced task-specific engineering and improved cross-task reuse.

Despite this promise, directly deploying VLA models in surgical settings reveals a critical conflict between their open-loop execution mechanism and the operational demands of surgical manipulation. Surgical manipulation operates at millimeter-level precision, imposing stringent constraints on trajectory smoothness and end-effector pose consistency~\cite{lee2025vaisi}. Besides, many procedures require sustained contact with soft tissue, whose nonlinear and viscoelastic characteristics complicate stable interaction~\cite{yang2025task}. Time-varying contact conditions induced by tissue retraction or instrument exchange further expose multi-step action predictions to out-of-distribution states~\cite{iftikhar2024artificial}. In practice, current VLA architectures typically execute fixed-length action chunks in an open-loop manner. Under these unstable conditions, committing to such a rigid open-loop horizon proves dangerous: continuing to execute blind, pre-planned actions when the target tissue unexpectedly deforms, slips, or alters the visual scene can directly lead to severe anatomical damage.

To mitigate the risks of open-loop execution during the automated execution of repetitive surgical subtasks, we propose the Trajectory Divergence Horizon Decision (TDHD) mechanism as a safety gate. Rather than relying on explicit modeling of complex contact dynamics, TDHD functions as an adaptive test-time monitor that evaluates the step-wise prediction reliability of a generated action chunk. By detecting underlying trajectory instability, this mechanism identifies the precise moment when continuing blind open-loop execution becomes hazardous. Consequently, the safety gate truncates the execution horizon at the first unstable step, effectively forcing the system to replan from updated visual observations before spatial errors can amplify. This targeted intervention helps the automated framework handle both stringent rigid-body constraints and unpredictable tissue deformations during execution.

We validate our automated framework on a real-world da Vinci-like dual-arm surgical robotic platform equipped with multi-view perception (external and wrist-mounted cameras), utilizing teleoperated demonstrations to fine-tune the policy. To explicitly test the system, we design two complementary task suites: needle manipulation (reach, pick, regrasp) to evaluate millimeter-level absolute precision under rigid-body limitations, and soft-tissue manipulation (reach, lift, resection) to assess adaptive truncation capabilities in the presence of nonlinear deformations and dynamic contact forces. Across these six hierarchical subtasks, TDHD consistently enhances execution reliability over strong VLA baselines, yielding higher end-to-end success rates in both task suites. The core contributions of this work are threefold:
\begin{itemize}
    \item We establish a real-world dual-arm surgical manipulation setup to advance the automated execution of repetitive subtasks using Vision-Language-Action (VLA) models. This environment incorporates teleoperated demonstrations with multi-view perception and language instructions to evaluate six hierarchical subtasks that explicitly contrast rigid-object precision with deformable-tissue interactions.
    
    \item We propose the TDHD mechanism as a dynamic safety gate for this automated framework. As a targeted test-time strategy, TDHD evaluates trajectory instability via spatial divergence under minor noise perturbations in deterministic flow-matching generation, effectively mitigating the physical risks of open-loop execution.

    \item We conduct comprehensive real-world evaluations demonstrating improved execution reliability and stage-wise success rates compared with recent VLA-based policies on dual-arm surgical manipulation tasks, achieving end-to-end success rates of 60\% and 80\% in needle and tissue manipulation respectively.
    \end{itemize}

\section{Related Work}

\subsection{Surgical Robot Autonomy and Learning-based Control}
Surgical robot autonomy has advanced from scripted, geometry-driven behaviors~\cite{murali2015learning, alterovitz2003sensorless} toward perception-aware decision-making, adaptive manipulation, and task-level reasoning in complex and unstructured environments~\cite{tang2026geolang,huang2026mosformer,schmidgall2024general,zheng2024user}. Recent studies have explored autonomous subtasks such as tissue handling and retraction using learning-based paradigms, including demonstration-guided reinforcement learning and related formulations~\cite{singh2023autonomous}. To mitigate the high cost and safety constraints of real-world exploration, dedicated simulation platforms for surgical robotics have been developed, with sim-to-real transfer used to validate learned policies under physical deployment~\cite{xu2021surrol}. 

Learning from demonstration and imitation learning remain the dominant approaches for surgical skill acquisition~\cite{kim2024surgical,mazza2026moe}. By using expert demonstrations, these methods can encode complex procedural knowledge into control policies~\cite{moghani2025sufia}. Nevertheless, surgical manipulation poses several challenges: limited annotated data, the constrained and coupled kinematics of surgical manipulators, and stringent safety requirements that demand predictable and stable behavior~\cite{wang2025copesd}. To address long-horizon execution and error accumulation, hierarchical control frameworks have recently been proposed to decompose procedures into structured subtasks, thereby improving robustness and generalization in realistic surgical settings~\cite{kim2025srt}. In addition, Long et al.~\cite{long2025surgical} presented an open-source framework for surgical embodied intelligence with the goal of enabling generalized task autonomy beyond narrowly defined environments.

\subsection{Vision-Language-Action Models}

Compared to prior learning-based methods that are typically used to train separate, task-specific policies, Vision-Language-Action (VLA) models shift the locus of generalization into a single language-conditioned policy by unifying multimodal perception, semantic reasoning, and action generation within end-to-end architectures, and have emerged as a promising paradigm for general-purpose robotic intelligence~\cite{sapkota2025vision}. RT-2~\cite{zitkovich2023rt} extended large vision-language models to robotics by casting low-level actions as text tokens, which facilitates knowledge transfer from web-scale pretraining to embodied control. In parallel, policy-centric approaches such as Octo~\cite{team2024octo} and OpenVLA~\cite{kim2024openvla} trained transformer-based policies on large and heterogeneous robot datasets, emphasizing cross-task and cross-embodiment generalization. Subsequent foundation policies, including $\pi_0$~\cite{black2024pi_0}, $\pi_{0.5}$~\cite{intelligence2025pi_}, and G0~\cite{jiang2025galaxea}, further scaled model capacity and data diversity, demonstrating improved robustness and broader skill coverage. Beyond autoregressive transformers, RDT~\cite{liu2024rdt} introduced a diffusion-based formulation that generates temporally consistent action trajectories, which is beneficial for long-horizon and multimodal control.

Recent efforts have begun to investigate medical adaptations of this paradigm. RoboNurse-VLA~\cite{li2025robonurse} proposed language-guided assistive policies for clinical support tasks with explicit safety considerations. EndoVLA~\cite{ng2025endovla} explored multimodal grounding in endoscopic environments and demonstrated the feasibility of aligning perception and action in minimally invasive scenarios. However, existing approaches primarily emphasize single-arm manipulation or perception-centric formulations, and coordinated multi-instrument surgical manipulation remains underexplored.

\section{Method}

\begin{figure*}[h]
    \centering
    % \includesvg[width=\textwidth]{system_overview.pdf}
    \includegraphics[width=\textwidth]{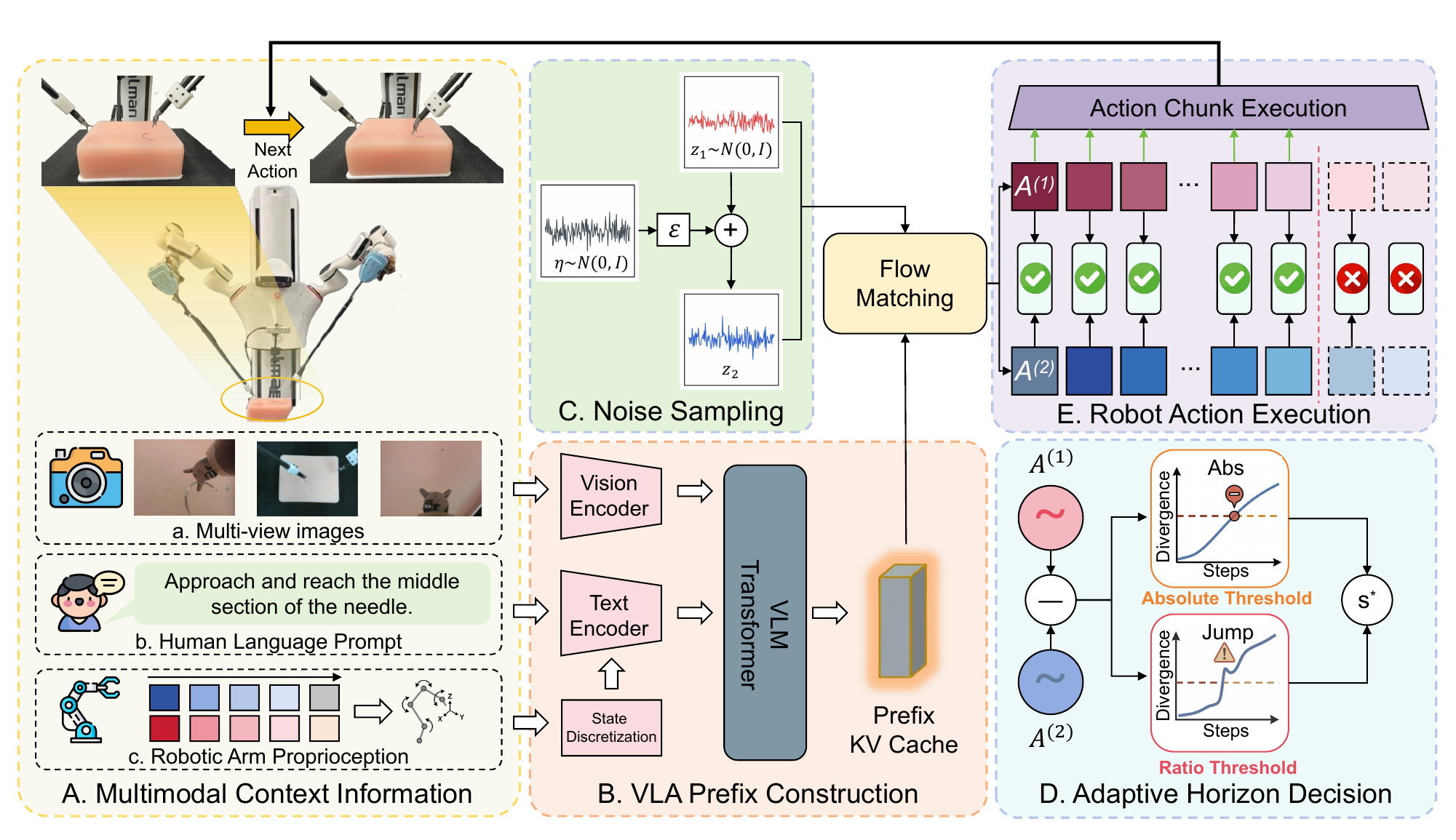}
\caption{\textbf{Overview of the automated surgical manipulation framework with TDHD.} 
(A)~Multimodal context information includes multi-view images, a language instruction, and robotic arm proprioception; 
(B)~VLA prefix construction, where visual, linguistic, and proprioceptive tokens are fused by a VLM Transformer into a Prefix KV Cache computed once per cycle; 
(C)~Noise sampling creates a primary noise $z_1$ and a perturbed copy $z_2{=}z_1{+}\epsilon\eta$; 
(D)~Adaptive horizon decision, where per-step divergence $\delta_i$ between trajectories $A^{(1)}$ and $A^{(2)}$ is evaluated against absolute and ratio thresholds to determine the safe execution length $s^*$; 
(E)~Robot action execution, where only the first $s^*$ reliable steps of $A^{(1)}$ are dispatched to the robot.}
    \label{fig:system_overview}
\end{figure*}

\subsection{VLA Model With Action Chunking}
A Vision-Language-Action policy acts as the core controller for the robot. At any given time step $t$, the VLA model receives a current observation $o_t$ (which includes camera images and robot joint states) and a language instruction $l$. Instead of predicting only the next single action, the model uses an action chunking strategy that predicts a complete sequence of future actions all at once. We define this output sequence as an action chunk $A$. If the model predicts a fixed length of $H$ steps, the action chunk is written as $A = [a_t, a_{t+1}, \dots, a_{t+H-1}]$. Each individual action $a_i$ contains the target positions for the robot arms and the instruments. By predicting this $H$-step sequence $\pi(A | o_t, l)$, the model helps the robot perform smooth and continuous movements.

\subsection{Flow Matching}
To efficiently generate the action chunk $A$, the VLA system uses a two-part network structure with flow matching. A large encoder network processes the images and instructions, caching the resulting representations. Given its computational cost, the system only runs it once per chunk. Then, a much smaller decoder network takes over to predict the actual robotic motions. Flow matching treats the generation as a continuous path from pure noise to a valid action sequence. At time $\tau=1$, the state is just a random noise matrix $z \sim \mathcal{N}(0, I)$. The decoder predicts the movement direction $v_\theta(x_\tau, \tau, c)$ at each moment. By moving along this direction from $\tau=1$ to $\tau=0$, the system removes the noise and arrives at the final action block $A$. The formula is:
\begin{equation}
A = z + \int_{1}^{0} v_\theta(x_\tau, \tau, c) \mathrm{d}\tau.
\end{equation}
Since this mathematical path is completely deterministic, there is no extra randomness during the steps. Given fixed conditioning context $c$, model parameters $\theta$, and ODE solver configuration, the generated action chunk $A$ is fully determined by the initial noise sample $z$.

\subsection{Trajectory Divergence Horizon Decision}
Existing VLA models typically execute a fixed number of steps $s$ from the predicted action chunk $A$. However, the reliability of the predicted actions tends to drop as the step index grows. When the robot performs complex tasks like inserting or cutting, the surgical environment introduces nonlinear tissue responses and unmodeled contact perturbations. As the prediction horizon extends, the model faces multiple possible correct movements, causing the predicted action distributions to become multimodal and spatially diffuse. This creates an inherent trade-off when $s$ is fixed: choosing a large $s$ better exploits the model's long-range intent but increases open-loop exposure when later actions drift from the true system evolution, whereas choosing a small $s$ limits execution to near-term, lower-uncertainty steps and can introduce discontinuities at chunk boundaries.

To address this challenge, as illustrated in Fig.~\ref{fig:system_overview}, we introduce the TDHD mechanism as a test-time safety gate that adaptively determines the reliable execution horizon of the generated action chunk based on trajectory divergence. Since the flow matching ODE (Ordinary Differential Equation) is fundamentally deterministic, the final action trajectory $A$ is entirely conditioned on the initial noise sample $z$. Motivated by local Lyapunov exponents, which characterize dynamical stability via the divergence of nearby trajectories~\cite{eckhardt1993local}, we treat the sensitivity of the generated trajectory to initial noise perturbations as a practical proxy for trajectory instability, similar to recent VLA frameworks that filter sampled actions via outcome verification~\cite{wu2025you}. In our surgical manipulation setting, this divergence reflects both epistemic uncertainty from under-represented states and aleatoric uncertainty from stochastic deformable tissue contact~\cite{chua2018deep}. Consequently, significant divergence indicates that the predicted sequence has become unstable, thereby reducing the reliability of subsequent open-loop execution.

To quantify this divergence for horizon decision, as depicted in the noise sampling module of Fig.~\ref{fig:system_overview}, we sample a primary noise vector $z_1 \sim \mathcal{N}(0, I)$ and construct a perturbed counterpart $z_2 = z_1 + \epsilon \eta$ during each inference pass, where $\eta \sim \mathcal{N}(0, I)$ denotes a random direction and $\epsilon$ is a small scaling factor. Both $z_1$ and $z_2$ are processed through the identical learned velocity field $v_\theta$ via flow matching, yielding two distinct action trajectories, $A^{(1)}$ and $A^{(2)}$. Given that the action space is normalized, we employ the $L_2$ distance to measure the step-wise trajectory divergence without requiring additional dimensional weighting. The divergence at each future step $i$ is formulated as:
\begin{equation}
\delta_i = \| A^{(1)}_i - A^{(2)}_i \|_2, \quad i \in \{0, 1, \dots, H-1\}.
\end{equation}

A smaller $\delta_i$ indicates higher stability in the predicted action at step $i$, making it safe for open-loop execution (denoted by the green checkmarks in the Action Chunk Execution module). Conversely, a larger $\delta_i$ indicates higher sensitivity to the noise initialization, suggesting reduced prediction reliability and that the action should be discarded (denoted by the red cross marks). To determine the dynamic execution horizon $s^*$, we stop the execution as soon as this divergence breaks the predefined stability limits. As shown in the adaptive horizon decision module of Fig.~\ref{fig:system_overview}, we set two independent stopping rules starting from step $i=1$:\\ 
\textbf{Absolute Threshold (Abs):} This detects significant differences using an absolute limit $\theta_{abs}$:
\begin{equation}
i_{abs} = \min \{ i \ge 1 \mid \delta_i > \theta_{abs} \}.
\end{equation}
\textbf{Ratio Threshold (Jump):} This identifies sharp changes in instability using a ratio limit $r_{thresh}$:
\begin{equation}
i_{rat} = \min \left\{ i \ge 1 \;\middle|\; \frac{\delta_i}{\delta_{i-1} + \xi} > r_{thresh} \right\},
\end{equation}
where $\xi$ is a very small positive constant to prevent dividing by zero. We determine the earliest step $\hat{s}$ at which either rule is broken:
\begin{equation}
\hat{s} = \min(i_{abs}, i_{rat}).
\end{equation}
If all steps are safe, $\hat{s}$ simply defaults to the maximum allowed length $s_{\max}$. Finally, to prevent motion discontinuities and excessive replanning, we constrain the final execution horizon to the physical range $[s_{\min}, s_{\max}]$, where $s_{\min}$ is a small positive lower bound:
\begin{equation}
s^* = \max \big( s_{\min}, \, \min ( \hat{s}, s_{\max} ) \big).
\end{equation}

After executing the selected $s^*$ actions, the automated system captures updated observations and predicts a new action chunk. Because the vision-language encoder is computed once per control cycle, evaluating the action decoder for both $z_1$ and $z_2$ introduces only modest additional computation. In repetitive surgical subtasks, prediction drift often accumulates during contact transitions, such as tissue traction and instrument exchange. TDHD mitigates this by truncating execution when divergence exceeds stability criteria, triggering immediate replanning to improve reliability under both rigid-body constraints and deformable-tissue interactions.

\label{sec:method}

\section{Experiment}
\subsection{Experimental Setup}

\subsubsection{Hardware Platform}
\begin{figure}[t]
    \centering
    \includegraphics[width=1\linewidth]{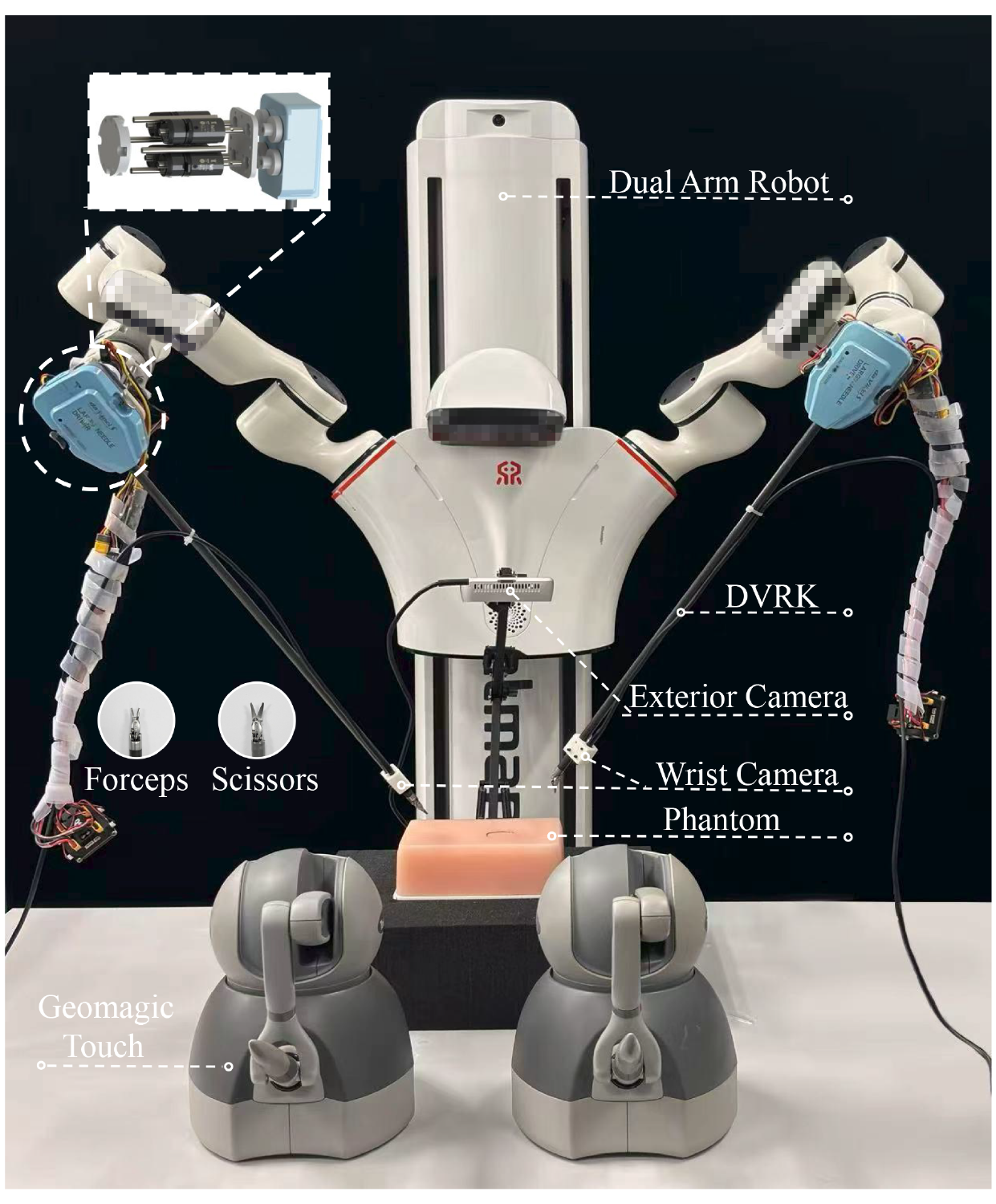}
    \caption{\textbf{Experimental setup of the dual-arm da Vinci-like surgical robot.} The platform integrates multi-view perception from exterior and wrist-mounted cameras and collects teleoperated demonstrations by controlling the dVRK via Geomagic Touch controllers.}
    \label{fig:exp_setup}
\end{figure}

Commercial robotic-assisted minimally invasive surgery (MIS) systems such as da Vinci\textsuperscript{\textregistered}~\cite{guthart2000intuitive}, Telelap Alf-X~\cite{altobelli20131405}, and MiroSurge~\cite{hagn2010dlr} are widely available. They provide interfaces for teleoperation while restricting low-level control access, resulting in a high barrier to entry and limited scalability for large-scale data collection and research exploration. Thus, we develop an accessible, da Vinci-like surgical robotic system built upon a realman RM65-B dual-arm robot, equipped with motorized surgical end-effectors adapted from the da Vinci Research Kit (dVRK), as illustrated in Fig.~\ref{fig:exp_setup}. Each manipulator provides large-range positioning capability, while dedicated distal motor units actuate the surgical instruments to enable precise grasping and cutting. A compliant soft phantom is placed at the center of the workspace to approximate anatomical tissue interaction under controlled and repeatable conditions.

The perception system integrates a global external RGB camera and wrist-mounted cameras located near the distal ends of both instruments. The exterior camera offers a workspace-level overview for monitoring and recording, whereas the wrist-mounted cameras capture first-person visual observations that closely resemble intraoperative endoscopic views. Teleoperated demonstrations are collected using two Geomagic Touch haptic devices, where master-side Cartesian poses are mapped to the slave manipulators with position scaling. Throughout data collection and evaluation, velocity limits and workspace constraints are strictly enforced to ensure safe operation.
\subsubsection{Task Definition}

\begin{figure*}[h]
    \centering
    \includegraphics[width=\textwidth]{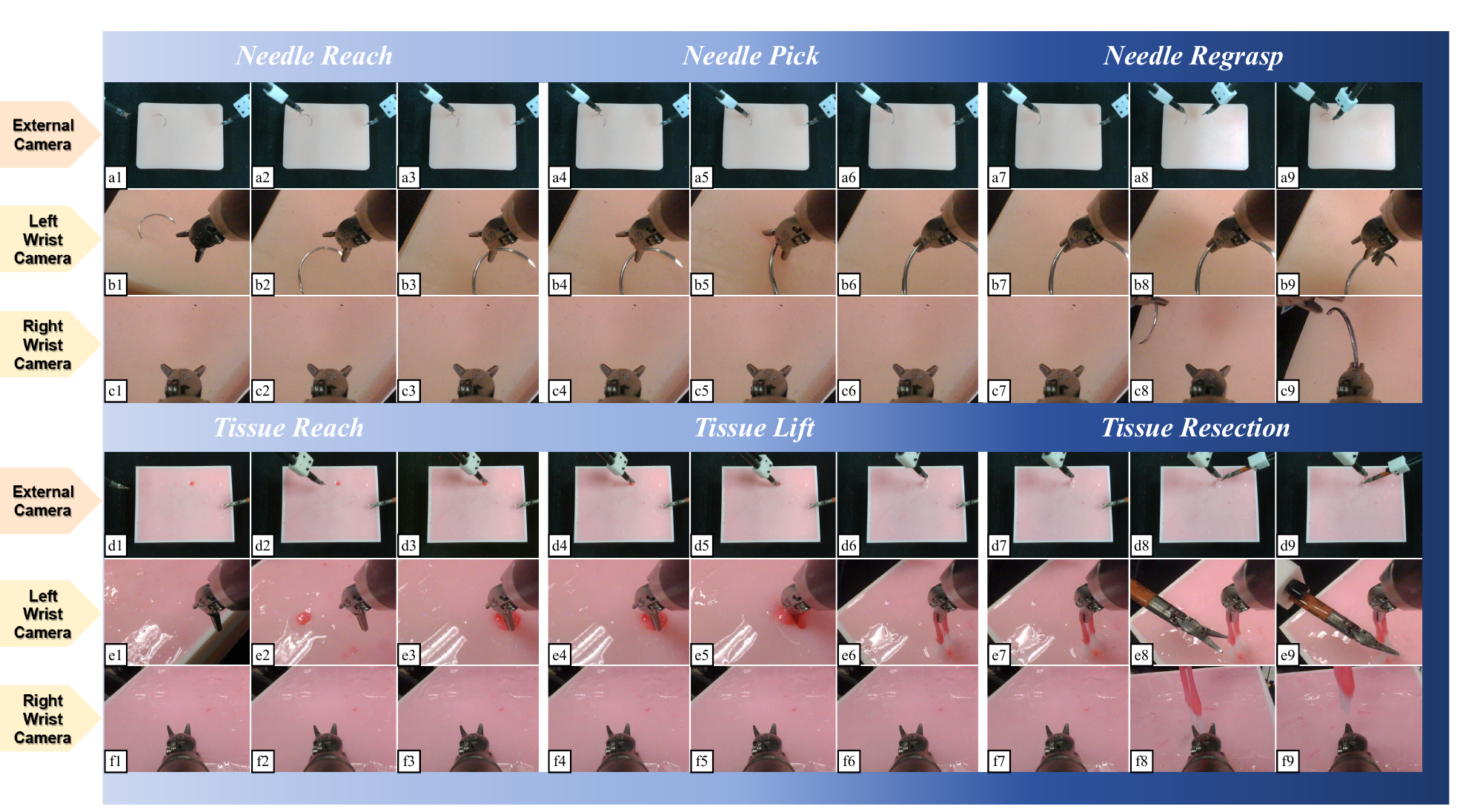}
    \caption{\textbf{Multi-view camera observations during real-world deployment.} Each column shows synchronized frames from three cameras (exterior, left-wrist, right-wrist) at the same timestamp. The horizontal axis represents the temporal progression of keyframes across needle manipulation (top) and tissue manipulation (bottom) task suites, illustrating the sequential reach, pick/lift, and regrasp/resection stages.}
    \label{fig:experiment result}
\end{figure*}

We design two task suites: needle manipulation and soft-tissue manipulation, covering distinct but representative skill axes in robot-assisted MIS (as shown in Fig.~\ref{fig:experiment result}). Needle manipulation instantiates tool–object precision handling primitives central to suturing workflows. Soft-tissue manipulation models tool-tissue interaction primitives common in lesion handling. Together, the two suites provide a balanced evaluation of (i) rigid-object precision and synchronization, and (ii) deformable-interaction stability, both of which are critical for reliable autonomous execution and directly probe the benefit of the proposed TDHD mechanism.

\textbf{Needle Manipulation Task Suite} follows a suturing-related sequence consisting of approach, grasp, and bimanual regrasp: (i) Needle Reach: The task requires positioning the tool center point (TCP) within $5\,\mathrm{mm}$ of the needle midpoint such that this midpoint lies inside the planar projection of the two gripper jaws onto the phantom surface. Satisfying both geometric conditions defines a reach success. (ii) Needle Pick: A trial is successful when the needle is securely grasped, lifted above the phantom surface without dropping, and held in its original curvature direction rather than flipped during grasping. (iii) Needle Regrasp: Following a successful pick, the needle is transferred from the left dVRK arm to the right dVRK arm. Regrasp success is defined by the right arm establishing a secure grasp, followed by release from the left arm, with the needle stably held throughout the transition.

\textbf{Tissue Manipulation Task Suite} models localized abnormality handling under deformable conditions within the same feasible workspace. The target abnormal point location varies across trials: (i) Tissue Reach: The TCP must reach the abnormal point such that the abnormal point lies within the planar projection of the two gripper jaws. These geometric constraints define reach success. (ii) Tissue Lift: After reaching, the abnormal region must be stably grasped and lifted above the phantom surface while maintaining grasp stability. Tissue deformation and elongation during lifting are permitted and do not constitute failure. (iii) Tissue Resection: The left arm maintains upward traction while the right arm cuts at the connection interface. Resection success requires complete severance of the abnormal region from the surrounding tissue such that the severed region can be further lifted without residual attachment.

\subsection{Dual-Arm Surgical Manipulation Dataset}

To build a solid basis for learning surgical control, we construct a real-world dataset based entirely on human demonstrations. Without using simulations, all tasks are performed via direct teleoperation using Geomagic Touch devices. This ensures accurate mapping from the human expert to the robot hardware. For each of the three subtasks (reach, pick/lift, regrasp/resection) in both the needle and tissue suites, we collected exactly 100 successful trials. Among these, we allocated 80 trials for the training set and 20 for testing. Consequently, each task suite comprises 300 episodes, resulting in a total dataset of 600 demonstrations.

To synchronize all hardware devices, the system uses a Network Time Protocol (NTP) global timestamping method at a strict 30 Hz frequency. The observation space combines visual data, robot states, and language text into a single HDF5 structure. Visual data are captured by the exterior camera for global RGB-D sensing and two wrist cameras for local RGB views, as shown in Fig.~\ref{fig:exp_setup}, all captured at a $640\times480$ resolution. Robot state data include joint positions from two 6-DoF manipulators, the 6-DoF end-effector poses of both arms, and the states and positions of the four gripper motors. Finally, each collected episode is paired with a natural language instruction to help the model understand and generalize the specific task.

\subsection{Evaluation Criteria}

\begin{table*}[htbp]
\centering
\caption{Quantitative Performance Success Rate for Two Task Suites (\%)}
\label{success rate}

\resizebox{\textwidth}{!}{
\begin{tabular}{l cccccc cccccc}
\toprule
\multirow{2}{*}{Method} & \multicolumn{6}{c}{Needle Manipulation} & \multicolumn{6}{c}{Tissue Manipulation} \\
\cmidrule(lr){2-7} \cmidrule(lr){8-13}
 & reach & pick & regrasp & total & $SR_{\mathrm{Pick}\mid\mathrm{Reach}}$ & $SR_{\mathrm{Regrasp}\mid\mathrm{Reach\ \&\ Pick}}$ & reach & lift & resection & total & $SR_{\mathrm{Lift}\mid\mathrm{Reach}}$ & $SR_{\mathrm{Resection}\mid\mathrm{Reach\ \&\ Lift}}$ \\
\midrule
$\pi_{0.5}$~\cite{intelligence2025pi_} & 100 & 85 & 55 & 55 & 85   & 64.7 & 80  & 80  & 55 & 55 & \textbf{100}  & 68.8 \\
$\pi_{0}$~\cite{black2024pi_0}   & 100 & 80 & 50 & 50 & 80   & 62.5 & 75  & 70  & 45 & 45 & 93.3 & 64.3 \\
RDT~\cite{liu2024rdt}   & 60  & 35 & 10 & 10 & 58.3 & 28.6 & 50  & 45  & 20 & 20 & 90   & 44.4 \\
Octo~\cite{team2024octo}  & 40  & 10 & 0  & 0  & 25   & 0    & 40  & 35  & 5  & 5  & 87.5 & 14.3 \\
Ours ($\pi_{0.5}$ + TDHD)  & 100 & 90& 60 & \textbf{60} & \textbf{90}   & \textbf{66.7} & 100 & 100 & 80 & \textbf{80} & \textbf{100}  & \textbf{80}   \\
\bottomrule
\end{tabular}}
\end{table*}

Since the tasks follow sequential dependencies, success is defined hierarchically. Let $S_1$, $S_2$, and $S_3$ denote reach, pick or lift, and final-stage success, respectively.
The end-to-end success rate and conditional success rates are defined as
\begin{equation}
SR_{e2e} = P(S_1 \cap S_2 \cap S_3),
\end{equation}
\begin{equation}
SR_{\mathrm{Pick}\mid\mathrm{Reach}} = P(S_2 \mid S_1),
\end{equation}
\begin{equation}
SR_{\mathrm{Final}\mid\mathrm{Reach\&Pick}} = P(S_3 \mid S_1 \cap S_2),
\end{equation}
which isolate stage-level performance under upstream success conditions.

\subsection{Quantitative Results}

To quantitatively evaluate the competing policies, including $\pi_{0.5}$, $\pi_{0}$, RDT, Octo, and our proposed method, we conducted 20 independent trials for each task setting. Each policy was evaluated across all six subtasks, resulting in a total of 600 real-world robotic trials (5 policies × 6 subtasks × 20 trials). For a fair comparison, all baseline models were fine-tuned on our teleoperated demonstration dataset. Beyond reporting the end-to-end success rate, we further measured conditional success rates at each stage. 
%This step-wise evaluation enabled fine-grained analysis of failure modes and isolated the physical bottlenecks that arose during sequential manipulation.

For the needle manipulation tasks, our method achieves the highest overall success rate of 60\%. At the initial reach stage, generalist policies such as Octo and RDT frequently fail to localize the target precisely, whereas both $\pi$-series methods and ours achieve a 100\% reach rate. The subsequent pick stage introduces stricter physical constraints. Even when the gripper visually covers the needle, slight misalignment causes uneven contact forces during closure. Under open-loop fixed-step execution, the jaws close without corrective feedback, which often induces torsional disturbance and needle flipping. Our method actively mitigates this instability and achieves a 90\% conditional pick success rate ($SR_{\mathrm{Pick}\mid\mathrm{Reach}}$), compared to 85\% for $\pi_{0.5}$. Although the numerical gain over $\pi_{0.5}$ is modest, fixed-step execution frequently leaves the needle slightly tilted after grasping, creating a non-planar pose that complicates bimanual alignment. In contrast, execution regulated by the TDHD framework alleviates this effect and promotes more stable planar alignment, facilitating regrasp initialization.

The regrasp stage demands precise synchronization in both spatial position and vertical alignment. The receiving arm must match the grasp height and orientation of the holding arm before release, making the task highly sensitive to small pose errors. Consequently, success rates decrease across all methods at this stage. Our model maintains a 66.7\% conditional regrasp success rate ($SR_{\mathrm{Regrasp}\mid\mathrm{Reach\ \&\ Pick}}$), outperforming the baselines. This robustness stems from the TDHD framework: by injecting small perturbations into the generation process and monitoring the trajectory distance ($\delta_i$), the system detects early instability and truncates execution before error amplification, enabling corrective replanning and stable dual-arm coordination.

For the tissue manipulation tasks, the performance gain from the TDHD framework is more pronounced. Compared to the rigid needle, the lesion point exhibits limited color contrast with the surrounding phantom, making precise localization more sensitive to small visual errors. As a result, both $\pi_{0.5}$ and $\pi_{0}$ show reduced reach success rates (80\% and 75\%, respectively), whereas our method maintains 100\% reach performance. Unlike the needle tasks, successful reach in the tissue task directly implies successful grasping. Once the tool aligns with the target region, lifting primarily depends on maintaining an appropriate grasping height rather than precise geometric centering. Consequently, conditional lift success rates remain high across methods, and our model achieves a 100\% $SR_{\mathrm{Lift}\mid\mathrm{Reach}}$.

The primary difficulty lies in the resection stage. Cutting requires sustained tension and consistent alignment between the scissor trajectory and the connection boundary of the deformable tissue. Because the material continuously deforms under traction, the cutting boundary shifts dynamically during execution. Fixed-step open-loop execution accumulates these geometric deviations, causing misaligned cuts or incomplete separation. In contrast, dynamic step selection mitigates error propagation by truncating unstable segments and replanning from updated observations. This adaptive behavior substantially improves robustness under deformable interaction, leading to an 80\% conditional resection success rate ($SR_{\mathrm{Resection}\mid\mathrm{Reach\ \&\ Lift}}$), outperforming all baselines.

\subsection{Ablation Study}
To study the effect of adaptive execution-horizon selection, we conduct an ablation study using $\pi_{0.5}$ model fine-tuned on our dataset. For each task, success rates are evaluated over 10 repeated runs under identical conditions. We compare three execution strategies: (1) Fixed $s_{\max}$, (2) Fixed $s_{\mathrm{best}}$ selected from $s \in \{5,10,15,20,25\}$, and (3) $\pi_{0.5}$+TDHD (Ours), which dynamically determines execution length based on trajectory stability. The selected Fixed $s_{\mathrm{best}}$ is $s=25$ for needle tasks and $s=15$ for tissue tasks. Absolute success rates vary slightly from earlier reports due to minor setup changes (e.g., changes in camera viewpoints), while all comparisons remain fair and consistent.

As shown in Table~\ref{tab:ablation}, the gain from TDHD is task-dependent. On needle tasks, TDHD improves Pick success from 80\% to 90\%, while Reach remains 100\% and Regrasp remains 50\%. On tissue tasks, TDHD improves Resection success from 60\% (Best Fixed-$s$) to 70\%, with Reach and Lift both at 90\%. These results indicate that adaptive horizon control is most beneficial in stages with stronger contact instability (needle pick and tissue resection), rather than in already stable stages.
Under the adaptive TDHD strategy, the average executed step lengths are 21.15, 22.08, and 20.40 for reach, pick, and regrasp, respectively.

To analyze when the TDHD safety gate is activated during execution, we record the executed step length for each action chunk throughout testing. For illustration, we analyze a representative single trial, where the average executed step lengths for needle manipulation are 21.15 (reach), 22.08 (pick), and 20.40 (regrasp), while tissue manipulation shows 19.43 (reach), 23.76 (lift), and 17.13 (resection). For needle pick, TDHD is mainly triggered during the downward alignment phase, preventing excessive pushing that may cause needle tipping. After grasping, the upward lift executes close to the full 25-step horizon, explaining the relatively high pick-phase average (22.08). For tissue manipulation, the near full-horizon execution in lift (23.76) aligns with the strong conditional success rate $SR_{\mathrm{Lift}\mid\mathrm{Reach}}$ in Table~\ref{success rate}, indicating stable grasping once alignment is achieved. In contrast, the shorter execution length in resection (17.13) reflects more frequent replanning during cutting. Overall, TDHD tends to truncate execution during delicate contact transitions while allowing longer horizons in stable motion phases.

\begin{table}[t]
\vspace{0.9em}
\centering
\caption{Ablation Study on Execution Strategies (\%).}
\label{tab:ablation}
\begin{tabular}{lccc|ccc}
\toprule
& \multicolumn{3}{c|}{Needle} & \multicolumn{3}{c}{Tissue} \\
\cmidrule(lr){2-4}\cmidrule(lr){5-7}
Method & Reach & Pick & Regrasp & Reach & Lift & Resection \\
\midrule
Fixed $s_{\max}$        & 100 & 80 & 50 & 70 & 70 & 40 \\
Fixed $s_{\mathrm{best}}$              & 100 & 80 & 50 & 90 & 90 & 60 \\
Ours & 100 & 90 & 50 &90 & 90 & 70 \\
\bottomrule
\end{tabular}
\end{table}

\section{Conclusion}

This work studies reliable real-world deployment of Vision-Language-Action (VLA) policies for repetitive dual-arm surgical subtasks. While VLA models provide a unified framework for perception, language grounding, and action generation, fixed-horizon open-loop execution can accumulate errors under changing scene conditions. By incorporating our proposed Trajectory Divergence Horizon Decision as a test-time execution-horizon controller, the system can truncate unreliable segments and replan from updated observations, improving robustness on both needle and tissue task suites, with the most notable benefits appearing in later, contact-sensitive stages. Nevertheless, our evaluation focuses on controlled phantom environments and a limited set of subtasks, leaving open questions about performance under more diverse anatomies, longer procedures, and richer sources of uncertainty. Future work will investigate threshold calibration, the impact of trajectory ensemble size in TDHD, and its consistency across alternative VLA models.

%%%%%%%%%%%%%%%%%%%%%%%%%%%%%%%%%%%%%%%%%%%%%%%%%%%%%%%%%%%%%%%%%%%%%%%%%%%%%%%%

%%%%%%%%%%%%%%%%%%%%%%%%%%%%%%%%%%%%%%%%%%%%%%%%%%%%%%%%%%%%%%%%%%%%%%%%%%%%%%%%

%%%%%%%%%%%%%%%%%%%%%%%%%%%%%%%%%%%%%%%%%%%%%%%%%%%%%%%%%%%%%%%%%%%%%%%%%%%%%%%%
% \section*{APPENDIX}

\section*{ACKNOWLEDGMENT}
We would like to thank Mr. Shuang Wu from Theory Lab, Central Research Institute, 2012 Labs, Huawei Technologies Co. Ltd., Hong Kong SAR, 999077, China, and Mr. Yang Yang for their inspiring discussions and valuable guidance.

%%%%%%%%%%%%%%%%%%%%%%%%%%%%%%%%%%%%%%%%%%%%%%%%%%%%%%%%%%%%%%%%%%%%%%%%%%%%%%%%

\balance
\bibliographystyle{IEEEtran}
\bibliography{reference}

\end{document}